\documentclass{ecai2014}
\usepackage{times}
\usepackage{graphicx}
\usepackage{latexsym}
\usepackage{hyperref}

\usepackage{amsmath,amssymb}
\newtheorem{definition}{Definition}

\newtheorem{proposition}{Proposition}

\usepackage{enumitem}
\setlist{itemsep=0pt,topsep=3pt,parsep=0pt,partopsep=0pt}

\usepackage[]{todonotes}

\usepackage{ifthen}
\newcommand{\brifnotempty}[1]{\ifthenelse{\equal{#1}{}}{}{ \br{#1}}}
\newenvironment{lemma*}[2][]
	{\pagebreak[2] \par \noindent \textbf{Lemma~\ref{#2}\brifnotempty{#1}.}\it}{\par}
\newenvironment{theorem*}[2][]
	{\pagebreak[2] \par \noindent \textbf{Theorem~\ref{#2}\brifnotempty{#1}.}\it}{\par}
\newenvironment{proposition*}[2][]
	{\pagebreak[2] \par \noindent \textbf{Proposition~\ref{#2}\brifnotempty{#1}.}\it}{\par}
\newenvironment{corollary*}[2][]
	{\pagebreak[2] \par \noindent \textbf{Corollary~\ref{#2}\brifnotempty{#1}.}\it}{\par}

\newcommand{\tuple}[1] {\langle #1 \rangle}
\newcommand{\im}{\leftarrow}

\newcommand{\obs}{\mathcal{O}}

\newcommand{\pws}[1]{2^{#1}}
\newcommand{\seq}[1]{\langle #1 \rangle}
\newcommand{\rlb}{\sigma}

\newcommand{\hrlb}{\hrl[\rlb]}
\newcommand{\hrl}[1][\rl]{H(#1)}
\newcommand{\lpif}{\leftarrow}
\newcommand{\brl}[1][\rl]{B(#1)}

\newcommand{\brlb}{\brl[\rlb]}
\newcommand{\brs}{\mathit{br}}
\newcommand{\mymodels}{\mathrel\mid\joinrel=}
\newcommand{\ent}{\mymodels}
\newcommand{\nf}{not\,}

\newcommand{\lgc}{L}

\newcommand{\kbs}{\mathbf{KB}}

\newcommand{\bss}{\mathbf{BS}}

\newcommand{\acc}{\mathbf{ACC}}

\newcommand{\bs}{S}

\newcommand{\brnot}{\mathop{\mathbf{not}\,}}
\newcommand{\sbrlits}{B}
\newcommand{\brlit}{L}
\newcommand{\mcs}{M}

\newcommand{\ctx}{C}
\newcommand{\kbmcs}{\mathit{kb}}

\newcommand{\fkb}[1]{F_{#1}}

\newcommand{\kbc}{knowledge base configuration}

\newcommand{\indi}{i}

\newcommand{\emMCS}{\mbox{eMCS}}
\newcommand{\evolving}{evolving}

\newcommand{\nxt}{next}
\newcommand{\now}{now}

\begin{document}

\title{On Minimal Change in Evolving Multi-Context Systems\\ (Preliminary Report)}

\author{Ricardo Gon\c{c}alves \and Matthias Knorr \and Jo\~{a}o Leite
\institute{CENTRIA \& Departamento de Inform{\'a}tica, Faculdade Ci\^encias e Tecnologia,
Universidade Nova de Lisboa, email: rjrg@fct.unl.pt} }

\maketitle
\bibliographystyle{ecai2014}

\begin{abstract}
Managed Multi-Context Systems (mMCSs) provide a general framework for integrating knowledge represented in heterogeneous KR formalisms. However, mMCSs are essentially static as they were not designed to run in a dynamic scenario. Some recent approaches, among them evolving Multi-Context Systems (\emMCS s), extend mMCSs by allowing not only the ability to integrate knowledge represented in heterogeneous KR formalisms, but at the same time to both react to, and reason in the presence of commonly temporary dynamic observations, and evolve by incorporating new knowledge. 
The notion of minimal change is a central notion in dynamic scenarios, specially in those that admit several possible alternative evolutions. 
Since eMCSs combine heterogeneous KR formalisms, each of which may require different notions of minimal change, the study of minimal change in eMCSs is an interesting and highly non-trivial problem.
In this paper, we study the notion of minimal change in \emMCS s, and discuss some alternative minimal change criteria.
\end{abstract}

\section{Introduction}


Multi-Context Systems (MCSs) were introduced in~\cite{BrewkaE07}, building on the work in~\cite{GiunchigliaS94,RoelofsenS05}, to address the need for a general framework that integrates  knowledge bases expressed in heterogeneous KR formalisms.
Intuitively, instead of designing a unifying language (see e.g., \cite{GoncalvesA10,MotikR10}, and \cite{KnorrAH11} with its reasoner NoHR \cite{IvanovKL13a}) to which other languages could be translated, in an MCS the different formalisms and knowledge bases are considered as modules, and means are provided to model the flow of information between them (cf.~\cite{AlbertiGGLS11,HomolaKLS12,HybridJLC13} and references therein for further motivation on hybrid languages and their connection to MCSs). 

More specifically, an MCS consists of a set of contexts, each of which is a knowledge base in some KR formalism, such that each context can access information from the other contexts using so-called bridge rules. 
Such non-monotonic bridge rules add their heads to the context's knowledge base provided the queries (to other contexts) in their bodies are successful.

Managed Multi-Context Systems (mMCSs) were introduced in~\cite{BrewkaEFW11} to provide an extension of MCSs by allowing operations, other than simple addition, to be expressed in the heads of bridge rules. This allows mMCSs to properly deal with the problem of consistency management within contexts.

One recent challenge for KR languages is to shift from static application scenarios which assume a one-shot computation, usually triggered by a user query, to open and dynamic scenarios where there is a need to react and evolve in the presence of incoming information. Examples include EVOLP~\cite{AlferesBLP02}, Reactive ASP \cite{GebserGKS11,GebserGKOSS12}, C-SPARQL \cite{BarbieriBCVG10}, Ontology Streams \cite{LecueP13} and ETALIS \cite{AnicicRFS12}, to name only a few. 

Whereas mMCSs are quite general and flexible to address the problem of integration of different KR formalisms, they are essentially static in the sense that the contexts do not evolve to incorporate the changes in the dynamic scenarios.

In such scenarios, new knowledge and information is dynamically produced, often from 
several different sources -- for example a stream of raw data produced by some sensors, new ontological axioms written by some user, newly found exceptions to some general rule, etc. 

To address this issue, two recent frameworks, \evolving\ Multi-Context Systems (\emMCS s)~\cite{ecai2014} and reactive Multi-Context Systems (rMCSs)~\cite{Brewka13,Ellmauthaler13,BrewkaEcai2014} have been proposed sharing the broad motivation of designing general and flexible frameworks inheriting from \mbox{mMCSs} the ability to integrate and manage knowledge represented in heterogeneous KR formalisms, and at the same time be able to incorporate knowledge obtained from dynamic observations. 

In such dynamic scenarios, where systems can have alternative evolutions, it is desirable to have some criteria of minimal change to be able to 
compare the possible alternatives. This problem is particularly interesting and non-trivial in dynamic frameworks based on MCSs, not only because of the heterogeneity of KR frameworks that may exist in an MCS -- each of which may require different notions of minimal change --, but also because the evolution of such systems is based not only on the semantics, but also
on the evolution of the knowledge base of each context. 

In this paper, we study minimal change in \emMCS s, by presenting some minimal change criteria to be applied to the possible evolving equilibria of an \emMCS, and by discussing the relation between them.

The remainder of this paper is structured as follows. After presenting the main concepts regarding mMCSs, we introduce the framework of \emMCS s. Then, we present and study some minimal change criteria in \emMCS s.
We conclude with discussing related work and possible future directions.

\section{Multi-Context Systems}\label{sec:prelim}

In this section, we introduce the framework of multi-context systems (MCSs).
Following \cite{BrewkaE07}, a multi-context system (MCS) consists of a collection of components, each of which contains knowledge represented in some \emph{logic}, defined as a triple $\lgc = \tuple{\kbs, \bss, \acc}$ where $\kbs$ is the set of well-formed knowledge bases of $\lgc$, $\bss$ is the set of possible belief sets, and $\acc: \kbs \rightarrow \pws{\bss}$ is a function describing the semantics of $\lgc$ by assigning to each knowledge base a set of acceptable belief sets. 
We assume that each element of $\kbs$ and $\bss$ is a set, and we define $\fkb{}=\{s:s\in \kbmcs \wedge \kbmcs \in \kbs\}$. 

In addition to the knowledge base in each component, \emph{bridge rules} are used to interconnect the components, specifying what knowledge to assert in one component given certain beliefs held in the components of the MCS. 
Formally, for a sequence of logics $\lgc = \seq{\lgc_1, \dotsc, \lgc_n}$, an \emph{$\lgc_i$-bridge rule $\rlb$ over $\lgc$}, $1 \leq i \leq n$, is of the form 
\begin{align}\label{bridgeRule}
\hrlb \lpif \brlb 
\end{align}
where $\brlb$ is a set of \emph{bridge literals} of the form $(r\!:\!b)$ and of the form $\brnot (r\!:\!b)$, $1 \leq r \leq n$, with $b$ a belief formula of $\lgc_r$, and, for each $\kbmcs \in \kbs_i$, $\kbmcs \cup \{\hrlb\} \in \kbs_i$.  

Bridge rules in MCSs only allow adding information to the knowledge base of their corresponding context. Note that the condition $\kbmcs \cup \{\hrlb\} \in \kbs_i$ precisely guarantees that the resulting set is still a knowledge base of the context.
In~\cite{BrewkaEFW11}, an extension, called managed Multi-Context Systems (mMCSs), is introduced in order to allow other types of operations to be performed on a knowledge base.
For that purpose, each context of an mMCS is associated with a \emph{management base}, which is a set of operations that can be applied to the possible knowledge bases of that context. Given a management base $OP$ and a logic $\lgc$, let $OF=\{op(s): op\in OP \wedge s\in \fkb{}\}$ be the \emph{set of operational formulas} that can be built from $OP$ and $\lgc$. 
Each context of an mMCS gives semantics to operations in its management base using a \emph{management function} over a logic $\lgc$ and a management base $OP$, $mng:\pws{OF}\times \kbs\rightarrow (\pws{\kbs}\setminus \{\emptyset\})$, i.e., $mng(op,kb)$  is the (non-empty) set of possible knowledge bases that result from applying the operations in $op$ to the knowledge base $kb$. 
We assume that $mng(\emptyset,kb)=\{kb\}$. 
Now, $\lgc_i$-bridge rules for mMCSs are defined in the same way as for MCSs, except that $\hrlb$ is now an operational formula over $OP_i$ and $L_i$.

\begin{definition}
A \emph{managed Multi-Context System} (mMCS) is a sequence $M=\tuple{\ctx_1,\ldots,\ctx_n}$, where each $\ctx_i$, $i\in\{1,\ldots,n\}$, called a \emph{managed context}, is defined as $\ctx_i=\tuple{\lgc_\indi,\kbmcs_\indi,\brs_\indi,OP_\indi, mng_\indi}$ where $L_\indi=\tuple{\kbs_{\indi},\bss_{\indi},\acc_{\indi}}$ is a logic, $kb_\indi\in \kbs_{\indi}$, $\brs_\indi$ is a set of $\lgc_\indi$-bridge rules, $OP_\indi$ is a management base, $mng_\indi$ is a management function over $L_\indi$ and $OP_\indi$.
 \end{definition}
 
  Note that, for the sake of readability, we consider a slightly restricted version of mMCSs where each $\acc_{\indi}$ is still a function and not a set of functions as for logic suites \cite{BrewkaEFW11}.

For an mMCS $\mcs = \seq{\ctx_1, \dotsc, \ctx_n}$, a \emph{belief state of $\mcs$} is a sequence $\bs = \seq{\bs_1, \dotsc, \bs_n}$ such that each $\bs_i$ is an element of $\bss_{\indi}$. For a bridge literal $(r:b)$, $\bs\ent (r:b)$ if $b \in \bs_r$ and $\bs \ent \brnot (r:b)$ if $b \notin \bs_r$; for a set of bridge literals $\sbrlits$, $\bs \ent \sbrlits$ if $\bs \ent\brlit$ for every $\brlit \in \sbrlits$.
We say that a bridge rule $\rlb$ of a context $\ctx_i$ is \emph{applicable given a belief state $\bs$ of $\mcs$} if $\bs$ satisfies $\brlb$. 
We can then define $app_\indi(\bs)$, the set of heads of bridge rules of $\ctx_i$ which are applicable in $\bs$, by setting $app_\indi(\bs)=\{\hrlb:\rlb\in \brs_\indi \wedge \bs\ent \brlb\}$.

Equilibria are belief states that simultaneously assign an acceptable belief set to each context in the mMCS such that the applicable operational formulas in bridge rule heads are taken into account.
\begin{definition}
A belief state $\bs = \seq{\bs_1, \dotsc, \bs_n}$ of an mMCS $\mcs$ is an \emph{equilibrium} of $\mcs$ if, for every $1 \leq i \leq n$, \[\bs_i \in \acc_{\indi}(kb)\  \ \text{ for some }\ kb\in mng_\indi(app_\indi(\bs),\kbmcs_\indi).\] 
\end{definition}

\section{Evolving Multi-Context Systems}\label{sec:emMCS}

In this section, we recall \evolving\ Multi-Context Systems, which generalize mMCSs to a dynamic scenario in which contexts are enabled to react to external observations and evolve.
For that purpose, we consider that some of the contexts in the MCS become so-called \emph{observation contexts} whose knowledge bases will be constantly changing over time according to the observations made, similar, e.g., to streams of data from sensors.\footnote{For simplicity of presentation, we consider discrete steps in time here.}

The changing observations then will also affect the other contexts by means of the bridge rules.
As we will see, such effect can either be instantaneous and temporary, i.e., limited to the current time instant, similar to (static) mMCSs, where the body of a bridge rule is evaluated in a state that already includes the effects of the operation in its head, or persistent, but only affecting the next time instant. 
To achieve the latter, we extend the operational language with a unary meta-operation $\nxt$ that can only be applied on top of operations.

\begin{definition}
Given a management base $OP$ and a logic $\lgc$, we define  $eOF$, the evolving operational language, 
as \[eOF=OF\cup\{\nxt(op(s)):op(s)\in OF\}.\]
 \end{definition}

The idea of observation contexts is that each such context has a language describing the set of possible observations of that context, along with its current observation. The elements of the language of the observation contexts can then be used in the body of bridge rules to allow contexts to access the observations.
Formally, an \emph{observation context} is a tuple $O=\tuple{\Pi_{O},\pi}$ where $\Pi_{O}$ is the \emph{observation language} of $O$ and $\pi\subseteq \Pi_{O}$ is its \emph{current observation}.

We can now define \evolving\ Multi-Context Systems.

\begin{definition}
An \emph{evolving Multi-Context Systems} (\emMCS) is a sequence $M_e=\tuple{\ctx_1,\ldots,\ctx_n,O_1,\ldots,O_\ell}$, such that, for each $i\in\{1,\ldots,\ell\}$, $O_i=\tuple{\Pi_{O_i},\pi_i}$ is an \emph{observation context}, and, for each $i\in\{1,\ldots,n\}$, $\ctx_i$ is an \emph{\evolving\ context}  defined as $\ctx_i=\tuple{\lgc_\indi,\kbmcs_\indi,\brs_\indi,OP_\indi, mng_\indi}$ where \vspace{0.1cm}
\begin{itemize}
 \item $L_\indi=\tuple{\kbs_{\indi},\bss_{\indi},\acc_{\indi}}$ is a logic \vspace{0.1cm}
 
 \item $kb_\indi\in \kbs_{\indi}$\vspace{0.1cm}
 
 \item $\brs_\indi$ is a set of bridge rules of the form \vspace{0.1cm}
\begin{align}\label{evolvingBridgeRule}
\hrlb\lpif a_1,\ldots, a_k, \brnot a_{k+1}, \ldots, \brnot a_{n}
\end{align}
 such that $\hrlb\in eOF_\indi$, and each $a_\indi$, $i\in\{1,\ldots,n\}$, is either of the form $(r\!:\!b)$ with $r\in\{1,\ldots,n\}$ and $b$ a belief formula of $L_r$, or 
 of the form $(r@b)$ with $r\in\{1,\ldots,\ell\}$ and $b\in \Pi_{O_r}$ \vspace{0.1cm}

  \item $OP_\indi$ is a management base\vspace{0.1cm}
 
 \item $mng_\indi$ is a management function over $L_\indi$ and $OP_\indi$.
  
\end{itemize}
\end{definition}

Given an \emMCS\ $M_e=\tuple{\ctx_1,\ldots,\ctx_n,O_1,\ldots,O_\ell}$ we denote by $\kbs_{M_e}$ the set of \emph{\kbc s for $M_e$}, i.e., $\kbs_{M_e}=\{\tuple{k_1,\ldots,k_n}:   k_i\in \kbs_{\indi} \text{ for each } 1\leq i\leq n \}$. 

\begin{definition} Let $M_e=\tuple{\ctx_1,\ldots,\ctx_n,O_1,\ldots,O_\ell}$ be an \emMCS.
A \emph{belief state} for $M_e$ is a sequence $S=\tuple{S_{1},\ldots,S_{n}}$ such that, for each 
 $1\leq i\leq n$, we have $S_i\in \bss_{\indi}$. We denote by $\bss_{M_e}$ the set of belief states for $M_e$. 
\end{definition}

  An \emph{instant observation} for $M_e$ is a sequence $\obs=\tuple{o_1,\ldots, o_{\ell}}$ such that, for each $1\leq i \leq \ell$, we have that $o_{i}\subseteq  \Pi_{O_\indi}$.

 The notion $app_\indi(\bs)$ of the set of heads of bridge rules of $\ctx_i$ which are applicable in a belief state $\bs$, cannot be directly transferred from mMCSs to \emMCS s since bridge rule bodies can now contain atoms of the form $(r@b)$, whose satisfaction depends on the current observation. Formally, given a belief state $S=\seq{\bs_1, \dotsc, \bs_n}$ for $M_e$ and an instant observation $\obs=\tuple{o_1,\ldots,o_\ell}$ for $M_e$, we define the satisfaction of bridge literals of the form $(r\!:\!b)$ as $S,\obs\ent (r\!:\!b)$ if $b\in S_r$ and  $\bs,\obs \ent \brnot (r\!:\!b)$ if $b \notin S_r$. 
The satisfaction of bridge literal of the form $(r@b)$ depends on the current observations, i.e., we have that $\bs,\obs\ent (r@b)$ if $b \in o_r$ and $\bs \ent \brnot (r@b)$ if $b \notin o_r$. 
As before, for a set $\sbrlits$ of bridge literals, we have that $\bs,\obs \ent \sbrlits$ if $\bs,\obs \ent\brlit$ for every $\brlit \in \sbrlits$.

We say that a bridge rule $\rlb$ of a context $\ctx_i$ is \emph{applicable given a belief state $\bs$ and an instant observation $\obs$} if its body is satisfied by $S$ and $\obs$, i.e., $\bs,\obs\ent\brlb$. 

We denote by $app_\indi(S,\obs)$ the set of heads of bridges rules of the context $C_i$ which are applicable given the belief state $S$ and the instant observation $\obs$. In the case of \emMCS s, the set $app_\indi(S,\obs)$ may contain two types of formulas: those that contain $\nxt$ and those that do not. 
As already mentioned before, the former are to be applied to the current knowledge base and not persist, whereas the latter are to be applied in the next time instant and persist.

\begin{definition}
Let $M_e=\tuple{C_1,\ldots,C_n,O_1,\ldots,O_\ell}$ be an \emMCS, $S$ a belief state for $M_e$, and $\obs$ an instant observation for $M_e$.
Then, for each $1\leq i\leq n$, consider the following sets:\vspace{0.2cm}
\begin{itemize}

 \item  $app_\indi^{\nxt}(S,\obs)=\{op(s):\nxt(op(s))\in app_\indi(S,\obs)\}$\vspace{0.2cm}

\item  $app_\indi^{\now}(S,\obs)=\{op(s):op(s)\in app_\indi(S,\obs)\}$

\end{itemize}

\end{definition}

Note that if we want an effect to be instantaneous and persistent, then this can also be achieved using two bridge rules with identical body, one with and one without $\nxt$ in the head.

Similar to equilibria in mMCS, the (static) equilibrium is defined to incorporate instantaneous effects based on $app_\indi^{\now}(S,\obs)$ alone.

\begin{definition}\label{def:staticEquilibrium}
Let $M_e=\tuple{C_1,\ldots,C_n,O_1,\ldots,O_\ell}$  be an \emMCS, and $\obs$ an instant observation for $M_e$.
A belief state $S=\tuple{S_{1},\ldots,S_{n}}$  for $M_e$ is an \emph{equilibrium of $M_e$ given $\obs$}  iff 
for each $1\leq i\leq n$, there exists some $kb\in mng_\indi(app^{\now}_\indi(S,\obs),kb_\indi)$
such that  $S_{i}\in \acc_{\indi}(kb)$.
\end{definition}

To be able to assign meaning to an \emMCS\ evolving over time, we introduce evolving belief states, which are sequences of belief states, each referring to a subsequent time instant.
\begin{definition}
Let $M_e=\tuple{C_1,\ldots,C_n,O_1,\ldots,O_\ell}$ be an \emMCS.
 An \emph{evolving belief state} of size $s$ for $M_e$  is a sequence $S_e=\tuple{S^1,\ldots,S^s}$ where each $S^j$,  $1\leq j\leq s$,  is a belief state for $M_e$.
  \end{definition}

To enable an \emMCS\ to react to incoming observations and evolve, a sequence of observations defined in the following has to be processed.
The idea is that the knowledge bases of the observation contexts $O_i$ change according to that sequence.

\begin{definition}
Let $M_e=\tuple{C_1,\ldots,C_n,O_1,\ldots,O_\ell}$ be an \emMCS.
A \emph{sequence of observations} for $M_e$ is a sequence $Obs=\tuple{\obs^1,\ldots,\obs^m}$, such that, for each  $1\leq j \leq m$, $\obs^j=\tuple{o_1^j,\ldots, o_{\ell}^j}$ is an instant observation for $M_e$, i.e., $o_{i}^j\subseteq  \Pi_{O_\indi}$ for each $1\leq i \leq \ell$. 
\end{definition}

To be able to update the knowledge bases and the sets of bridge rules of the \evolving\ contexts, we need the following notation.
Given an \evolving\ context $\ctx_\indi$, and a knowledge base $k\in \kbs_{\indi}$, we denote by $\ctx_\indi[k]$ the \evolving\ context in which $kb_\indi$ is replaced by $k$, i.e., $\ctx_\indi[k]=\tuple{L_\indi,k,br_\indi,OP_\indi,mng_\indi}$.
For an observation context $O_i$, given a set $\pi\subseteq \Pi_{O_\indi}$ of observations for $O_\indi$, we denote by $O_\indi[\pi]$ the observation context in which its current observation is replaced by $\pi$, i.e., $O_\indi[\pi]=\tuple{\Pi_{O_\indi},\pi}$.
Given $K=\tuple{k_1,\ldots, k_n}\in \kbs_{M_e}$ a \kbc\ for $M_e$,
we denote by $M_e[K]$ the \emMCS\ $\tuple{C_1[k_1],\ldots,C_n[k_n],O_1,\ldots,O_\ell}$.

We can now define that certain evolving belief states are evolving equilibria of an $M_e=\tuple{\ctx_1,\ldots,\ctx_n,O_1,\ldots,O_\ell}$ given a sequence of observations $Obs=\tuple{\obs^1,\ldots,\obs^m}$ for $M_e$. 
The intuitive idea is that, given an evolving belief state $S_e=\tuple{S^1,\ldots,S^s}$ for $M_e$, in order to check if $S_e$ is an evolving equilibrium, we need to consider a sequence of \emMCS s, $M^1,\ldots, M^s$ (each with $\ell$ observation contexts), representing a possible evolution of $M_e$ according to the observations in $Obs$, such that each $S^j$ is a (static) equilibrium of $M^j$. 
The current observation of each observation context $O_i$ in $M^j$ is exactly its corresponding element $o^j_i$ in $\obs^j$.
For each evolving context $C_i$, its knowledge base in $M^j$ is obtained from the one in $M^{j-1}$ by applying the operations in $app_\indi^{next}(S^{j-1},\obs^{j-1})$.

\begin{definition}\label{def:evolvingEquilibrium} 
Let $M_e=\tuple{\ctx_1,\ldots,\ctx_n,O_1,\ldots,O_\ell}$ be an \emMCS, $S_e=\tuple{S^1,\ldots,S^s}$ an evolving belief state of size $s$ for $M_e$, and $Obs=\tuple{\obs^1,\ldots,\obs^m}$ an observation sequence for $M_e$ such that $m\geq s$. 
Then, $S_e$ is an \emph{evolving equilibrium} of size $s$ of $M_e$  given $Obs$ iff, for each $1\leq j \leq s$, $\bs^{j}$ is an equilibrium of 
$M^j=\tuple{C_{1}[k_{1}^{j}],\ldots,C_n[k^j_n],O_1[o^j_1], \ldots, O_\ell[o^j_\ell]}$
where, for each $1\leq i\leq n$,  $k_{i}^{j}$ is defined inductively as follows:\vspace{0.2cm}
\begin{itemize}
 \item $k_{i}^1=kb_\indi$\vspace{0.2cm}
 
 \item $k_{i}^{j+1}\in mng_\indi(app^{\nxt}(S^{j},\obs^{j}_i),k^j_i).$ 
 \end{itemize}
 
\end{definition}

We end this section by presenting a proposition about evolving equilibria that will be useful in the next section.
In Def.~\ref{def:evolvingEquilibrium}, the number of considered time instances of observations $m$ is greater or equal the size of the evolving belief state with the intuition that an equilibrium may also be defined for only a part of the observation sequence.
An immediate consequence is that any subsequence of an evolving equilibrium is an evolving equilibrium.

\begin{proposition}\label{prop:subsequenceEquilibrium}
 Let $M_e=\tuple{\ctx_1,\ldots,\ctx_n,O_1,\ldots,O_\ell}$  be an \emMCS\ and $Obs= \tuple{\obs^1,\ldots,\obs^m}$ an observation sequence for $M_e$.
 If $S_e=\tuple{S^1,\ldots,S^s}$ is an evolving equilibrium of size $s$ of $M_e$  given $Obs$, then, for each $1\leq j \leq s$, and every $j\leq k\leq m$, we have that $\tuple{S^1,\ldots,S^j}$ is an evolving equilibrium of size $j$ of $M_e$  given the observation sequence $\tuple{\obs^1,\ldots,\obs^k}$.
\end{proposition}

\section{Minimal change}


In this section, we discuss some alternatives for the notion of minimal change in \emMCS s. 
What makes this problem interesting is that in an \emMCS\ there are different parameters that we may want to minimize in a transition from one time instant to the next one. In the following discussion we focus on two we deem most relevant: the operations that can be applied to the knowledge bases, and the distance between consecutive belief states.

We start by studying minimal change at the level of the operations.
In the following discussion we consider fixed an \emMCS\ $M_e=\tuple{\ctx_1,\ldots,\ctx_n,O_1,\ldots,O_\ell}$. 

Recall from the definition of evolving equilibrium that, in the transition between consecutive time instants, the knowledge base of each context $C_\indi$ of $M_e$ changes according to the operations in $app_i^{\nxt}(S,\obs)$, and these depend on the belief state $S$ and the instant observation $\obs$.
The first idea to compare elements of this set of operations is to, for a fixed instant observation $\obs$, distinguish those equilibria of $M_e$ which generate a minimal set of operations to be applied to the current knowledge bases to obtain the knowledge bases of the next time instant.  
Formally, given a \kbc\ $K\in \kbs_{M_e}$ and an instant observation $\obs$ for $M_e$, we can define the set:
\begin{align*}
MinEq(K,\obs)=\{&S:  S \text{ is an equilibrium of } M_e[K] \text{ given } \obs\\
 &\text{and there is no equilibrium }S ' \text{ of } M_e[K] \text{ given } \obs\\
 &\text{such that, for all } i\in\{1,\ldots,n\} \text{ we have}\\
 & app_i^{\nxt}(S',\obs)\subset app_i^{\nxt}(S,\obs)\}
\end{align*}

This first idea of comparing equilibria based on inclusion of the sets of operations can, however, be too strict in most cases.
Moreover, different operations usually have different costs, and it may well be that, instead of minimizing based on set inclusion, we want to minimize the total cost of the operations to be applied. For that, we need to assume that each context has a cost function over the set of operations, i.e., $cost_i:OP_i\to \mathbb{N}$, where $cost_i(op)$ represents the cost of performing operation $op$.

Let $S$ be a belief state for $M_e$ and $\obs$ an instant observation for $M_e$. Then, for each $1\leq i\leq n$, we define the cost of the operations to be applied to obtain the knowledge base of the next time instant as:
\begin{align*}
Cost_\indi(S,\obs)=\sum_{op(s)\in app^{\nxt}_\indi(S,\obs)} cost_\indi(op)
\end{align*}

Summing up the cost for all \evolving\ contexts, we obtain the global cost of $S$ given $\obs$: 
\begin{align*}
Cost(S,\obs)=\sum_{i=1}^{n}Cost_i(S,\obs)
\end{align*}

Now that we have defined a cost function over belief states, we can define a minimization function over possible equilibria of \emMCS\ $M_e[K]$ for a fixed \kbc\ $K\in \kbs_{M_e}$.
Formally, given $\obs$ an instant observation for $M_e$,
we define the set of equilibria of $M_e[K]$ given $\obs$ which minimize the global cost of the operations to be applied to obtain the \kbc\ of the next time instant as:
\begin{align*}
MinCost(K,\obs)=&\{S: S \text{ is an equilibrium of } M_e[K] \text{ given } \obs \text{ and}\\
&\text{there is no equilibrium } S'\text{ of } M_e[K] \text{ given } \obs\\ 
&\text{such that } Cost(S',\obs)< Cost(S,\obs)\}
\end{align*}

Note that, instead of using a global cost, we could have also considered a more fine-grained criterion by comparing costs for each context individually, and define some order based on these comparisons.
Also note that the particular case of taking $cost_i(op)=1$ for every $i\in\{1,\ldots,n\}$ and every $op\in OP_i$, captures the scenario of minimizing the total number of operations to be applied.

The function $MinCost$ allows for the choice of those equilibria which are minimal with respect to the operations to be performed to the current \kbc\ in order to obtain the \kbc\ of the next time instant. Still, for each choice of an equilibrium $S$, we have to deal with the existence of several alternatives in the set $mng_i(app^{\nxt}_i(S,\obs),kb_i)$. 
Our aim now is to discuss how we can apply some notion of minimal change that allows us to compare the elements $mng_i(app^{\nxt}_i(S,\obs),kb_i)$. 
The intuitive idea is to compare the distance between the current equilibria and the possible equilibria resulting from the elements in $mng_i(app^{\nxt}_i(S,\obs),kb_i)$. Of course, given the possible heterogeneity of contexts in an \emMCS, we cannot assume a global notion of distance between belief sets. Therefore, we assume that each evolving context has its own distance function between its beliefs sets. Formally, for each $1\leq i\leq n$, we assume the existence of a distance function $d_i$, i.e., $d_i:\bss_i\times \bss_i\to \mathbb{R}$ satisfying for all $S_1,S_2,S_3\in \bss_i$ the conditions:\vspace{0.1cm}
\begin{enumerate}
 \item $d_i(S_1,S_2)\geq 0$ \vspace{0.1cm}
 \item $d_i(S_1,S_2)=0$\ iff\  $S_1=S_2$ \vspace{0.1cm}
 \item $d_i(S_1,S_2)=d_i(S_2,S_1)$ \vspace{0.1cm}
 \item $d_i(S_1,S_3)\leq d_i(S_1,S_2)+d_i(S_2,S_3)$ \vspace{0.1cm}
\end{enumerate}

There are some alternatives to extend the distance function of each context to a distance function between belief states.
In the following we present two natural choices.
One option is to consider the maximal distance between belief sets of each context. Another possibility is to consider the average of distances between belief sets of each context. Formally, given $S^1$ and $S^2$ two belief states of $M_e$ we define two functions $\overline{d}_{\text{max}}:\bss_{M_e}\times\bss_{M_e}\to \mathbb{R}$ and $\overline{d}_{\text{avg}}:\bss_{M_e}\times\bss_{M_e}\to \mathbb{R}$ as follows:\vspace{0.2cm}
\begin{align*}
 \overline{d}_{\text{max}}(S^1,S^2)&=Max\{d_i(S^1_i,S^2_i)\ | \ 1\leq i \leq n\}\\[0.3cm]
 \overline{d}_{\text{avg}}(S^1,S^2)&=\frac{\sum_{i=1}^{n} d_i(S^1_i,S^2_i)}{n}
\end{align*}

We can easily prove that both $\overline{d}_{\text{max}}$ and  $\overline{d}_{\text{avg}}$ are indeed distance functions between belief states.

\begin{proposition}
The functions $\overline{d}_{\text{max}}$ and  $\overline{d}_{\text{avg}}$ defined above are both distance functions, i.e., satisfy the axioms (1- 4).
\end{proposition}

We now study how we can use one of these distance functions between belief states to compare the possible alternatives in the sets $mng_i(app^{\nxt}_i(S,\obs),kb_i)$, for each $1\leq i\leq n$. Recall that the intuitive idea is to minimize the distance between the current belief state $S$ and the possible equilibria that each element of $mng_i(app^{\nxt}_i(S,\obs),kb_\indi)$ can give rise to.
We explore here two options, which differ on whether the minimization is global or local.
The idea of global minimization is to choose only those \kbc s $\tuple{k_1,\ldots,k_n}\in \kbs_{M_e}$ with $k_i\in mng_i(app^{\nxt}_i(S,\obs),kb_i)$, which guarantee minimal distance between the original belief state $S$ and the possible equilibria of the obtained \emMCS. 
The idea of local minimization is to consider all possible tuples $\tuple{k_1,\ldots,k_n}$ with $k_i\in mng_i(app^{\nxt}_i(S,\obs),kb_i)$, and only apply minimization for each such choice, i.e., for each such \kbc\ we only allow equilibria with minimal distance from the original belief state.

We first consider the case of pruning those tuples $\tuple{k_1,\ldots,k_n}$ with $k_i\in mng_i(app^{\nxt}_i(S,\obs),kb_i)$ which do not guarantee minimal change with respect to the original belief state. 
We start with an auxiliary function. Let $S$ be a belief state for $M_e$, $K=\tuple{k_1,\ldots,k_n}\in \kbs_{M_e}$ a \kbc\ for $M_e$, and $\obs=\tuple{o_1,\ldots,o_\ell}$ an instant observation for $M_e$. Then we can define the set of \kbc s that are obtained from $K$ given the belief state $S$ and the instant observation $\obs$.
\begin{align*}
NextKB(S,\obs,\tuple{k_1,\ldots,k_n})=&\{\tuple{k'_1,\ldots,k'_n}\in \kbs_{M_e}:\\ 
&\text{for each }1\leq i\leq n, \text{ we have that}\\
& k'_i\in mng_i(app_i^{\nxt}(S,\obs),k_i) \}
\end{align*}

For each choice $\overline{d}$ of a distance function between belief states, we can define the set of \kbc s which give rise to equilibria of $M_e$ which minimize the distance to the original belief state. Let $S$ be a belief state for $M_e$, $K=\tuple{k_1,\ldots,k_n}\in \kbs_{M_e}$ a \kbc\ for $M_e$, and $\obs^j$ and $\obs^{j+1}$ instant observations for $M_e$.
\begin{align*}
MinNext(S,\obs^j,\obs^{j+1},&K)=\{(S',K'):\\
& K'\in NextKB(S,\obs^j,K) \text{ and } \\
&S'\in MinCost(M_e[K'],\obs^{j+1})\\
&\text{such  that there is no }\\
&K''\in NextKB(S,\obs^j,K) \text{ and no}\\
&S''\in MinCost(M_e[K''],\obs^{j+1})\\
&\text{with } \overline{d}(S,S'')<\overline{d}(S,S') \}.
\end{align*}

Again note that $MinNext$ applies minimization over all possible equilibria resulting from every element of $NextKB(S,\obs^j,K)$.
Using $MinNext$, we can now define a minimal change criterion to be applied to evolving equilibria of $M_e$. 

\begin{definition}
Let $M_e=\tuple{\ctx_1,\ldots,\ctx_n,O_1,\ldots,O_\ell}$ be an \emMCS, $Obs=\tuple{\obs^1,\ldots,O^m}$ an observation sequence for $M_e$, and  
let $S_e=\tuple{S^1,\ldots,S^s}$ be an evolving equilibrium of $M_e$ given $Obs$. We assume that $\tuple{K^1,\ldots,K^s}$, with $K^j=\tuple{k^j_1,\ldots,k^j_n}$, is the sequence of \kbc s associated with $S_e$ as in Definition~\ref{def:evolvingEquilibrium}.
Then, $S_e$ satisfies the \emph{strong minimal change criterion} for $M_e$ given $Obs$ if, for each $1\leq j\leq s$, the following conditions are satisfied: \vspace{0.2cm}
\begin{itemize}
 \item $S^j\in MinCost(M_e[K^j],\obs^{j})$\\
 
 \item $(S^{j+1},K^{j+1})\in MinNext(S^j,\obs^{j},\obs^{j+1},K^j)$
 
\end{itemize}

\end{definition}

We call this minimal change criterion the \emph{strong} minimal change criterion because it applies minimization over all possible equilibria resulting from every possible \kbc\ in $NextKB(S,\obs^j,K)$.

The following proposition states the desirable property that the existence of an equilibrium guarantees the existence of an equilibrium satisfying the strong minimal change criterion. We should note that this is not a trivial statement since we are combining minimization of two different elements: the cost of the operations and the distance between belief states. 
This proposition in fact follows from their careful combination in the definition of $MinNext$.

\begin{proposition}
Let $Obs=\tuple{\obs^1,\ldots,O^m}$ be an observation sequence for $M_e$. If $M_e$ has an evolving equilibrium of size $s$ given $Obs$, then there is at least one evolving equilibrium of size $s$ given $Obs$ satisfying the strong minimal change criterion.
\end{proposition}

Note that in the definition of the strong minimal change criterion, the \kbc s $K\in NextKB(S^j,\obs^j,K^j)$, for which the corresponding possible equilibria are not at a minimal distance from $S^j$, are not considered. 
However, there could be situations in which this minimization criterion is too strong. For example, it may well be that all possible \kbc s in $NextKB(S^j,\obs^j,K^j)$ are important, and we do not want to disregard any of them. 
In that case, we can relax the minimization condition by applying minimization individually for each \kbc\ in $NextKB(S^j,\obs^j,K^j)$. The idea is that, for each fixed $K\in NextKB(S^j,\obs^j,K^j)$ we choose only those equilibria of $M_e[K]$ which minimize the distance to $S^j$.

Formally, let $S$ be a belief state for $M_e$, $K\in \kbs_{M_e}$ a \kbc\ for $M_e$, and $\obs$ an instant observation for $M_e$. For each distance function $\overline{d}$ between belief states, we can define the following set:
\begin{align*}
MinDist(S,\obs,K)=&\{S': S'\in MinCost(M_e[K],\obs) \text{ and}\\
&\text{there is no } S''\in MinCost(M_e[K],\obs)\\
&\text{such that }\overline{d}(S,S'')< \overline{d}(S,S')\}
\end{align*}

Using this more relaxed notion of minimization we can define an alternative weaker criterion for minimal change to be applied to evolving equilibria of an \emMCS.

\begin{definition}
Let $M_e=\tuple{\ctx_1,\ldots,\ctx_n,O_1,\ldots,O_\ell}$ be an \emMCS, and $Obs=\tuple{\obs^1,\ldots,O^m}$ be an observation sequence for $M_e$, and  
 $S_e=\tuple{S^1,\ldots,S^s}$ be an evolving equilibrium of $M_e$ given $Obs$. We assume that $\tuple{K^1,\ldots,K^s}$, with $K^j=\tuple{k^j_1,\ldots,k^j_n}$, is the sequence of \kbc s associated with $S_e$ as in Definition~\ref{def:evolvingEquilibrium}.
Then, $S_e$ satisfies the \emph{weak minimal change criterion} of $M_e$ given $Obs$, if for each $1\leq j\leq s$ the following conditions are satisfied: \vspace{0.2cm}
\begin{itemize}
 \item $S^j\in MinCost(M_e[K^j],\obs^j)$\\
 
 \item $S^{j+1}\in MinDist(S^j,K^{j+1},\obs^{j+1})$
 
\end{itemize}

\end{definition}

In this case we can also prove that the existence of an evolving equilibrium implies the existence of an equilibrium satisfying the weak minimal change criterion. Again we note that the careful combination of the two minimizations -- cost and distance -- in the definition of $MinDist$, is fundamental to obtain the following result.

\begin{proposition}
Let $Obs=\tuple{\obs^1,\ldots,O^m}$ be an observation sequence for $M_e$. If $M_e$ has an evolving equilibrium of size $s$ given $Obs$, then 
$M_e$ has at least one evolving equilibrium of size $s$ given $Obs$ satisfying the weak minimal change criterion.
\end{proposition}

We can easily prove that, in fact, the strong minimal change criterion is stronger than the weak minimal change criterion.

\begin{proposition}
 Let $M_e=\tuple{\ctx_1,\ldots,\ctx_n,O_1,\ldots,O_\ell}$ be an \emMCS, and $Obs=\tuple{\obs^1,\ldots,O^m}$ be an observation sequence for $M_e$, and  
 $S_e=\tuple{S^1,\ldots,S^s}$ be an evolving equilibrium of $M_e$ given $Obs$. If $S_e$ satisfies the strong minimal change criterion of $M_e$ given $Obs$, then $S_e$ satisfies the weak minimal change criterion of $M_e$ given $Obs$.
\end{proposition}

We end this section by briefly discussing a global alternative to the minimization of costs of operations.
Recall that both $MinNext$ and $MinDist$ combine minimization of distances between belief states, and minimization of costs of operations. The minimization on costs is done at each time instant. Another possibility is, instead of minimizing the cost at each time instant $j$, to minimize the global cost of an evolving equilibrium.

We first extend the cost function over belief states to a cost function over evolving belief states. Let $S_e=\tuple{S^1,\ldots,S^s}$ be an evolving belief state for $M_e$, and $Obs=\tuple{\obs^1,\ldots,\obs^m}$ an observation sequence for $M_e$ such that $m\geq s$. Then we can define the cost of $S_e$ given $Obs$:
\begin{align*}
Cost(S_e,Obs)=\sum_{j=1}^{s}Cost(S^j,\obs^j)
\end{align*}

Let $Obs=\tuple{\obs^1,\ldots,\obs^m}$ be an observation sequence for $M_e$ and let $s\leq m$. We now define the set of evolving equilibria of size $s$ of $M_e$ given $Obs$ which minimize the total cost of the operations generated by it.
\begin{align*}
Min&Cost(M_e,Obs,s)=\{S_e: \\
&S_e \text{ is an evolving equilibrium of size }s \text{ of } M_e \text{ given } Obs \text{ and}\\
&\text{there is no evolving equilibrium } S'_e \text{ of size }s \text{ of } M_e \text{ given } Obs \\ 
&\text{such that } Cost(S'_e,Obs)< Cost(S_e,Obs)\}
\end{align*}

Since the minimization on costs of operations is orthogonal to the minimization on distances between belief states, it would be straightforward to use $MinCost(M_e,Obs,s)$ to define the global cost versions of $MinNext$ and $MinDist$, and use them to define the respective strong and weak minimal change criteria.

\section{Related and Future Work}\label{sec:conclusions}


In this paper we have studied the notion of minimal change in the context of the dynamic framework of \evolving\ Multi-Context Systems (\emMCS s)~\cite{ecai2014}.
We have presented and discussed some alternative definitions of minimal change criteria for evolving equilibria of a \emMCS.

Closely related to \emMCS s is the framework of reactive Multi-Context Systems (rMCSs)~\cite{Brewka13,Ellmauthaler13,BrewkaEcai2014} inasmuch as both aim at extending mMCSs to cope with dynamic observations. 
Three main differences distinguish them. 

First, whereas \emMCS s rely on a sequence of observations, each independent from the previous ones, rMCSs encode such sequences within the same observation contexts, with its elements being explicitly timestamped. This means that with rMCSs it is perhaps easier to write bridge rules that refer, e.g., to specific sequences of observations, which in \emMCS s would require explicit timestamps and storing the observations in some context, although at the cost that rMCSs need to deal with explicit time which adds an additional overhead. Second, since in rMCSs the contexts that result from the application of the management operations are the ones that are used in the subsequent state, difficulties may arise in separating non-persistent and persistent effects, for example, allowing an observation to override some fact in some context while the observation holds, but without changing the context itself -- such separation is easily encodable in \emMCS s given the two kinds of bridge rules, i.e., with or without the $\nxt$ operator. Finally, the bridge rules with the $\nxt$ operator allow the specification of transitions based on the current state, such as for example one encoded by the rule $\nxt(add(p))\im \nf p$, which do not seem possible in rMCSs. Overall, these differences seem to indicate that an interesting future direction would be to merge both approaches, exploring a combination of explicitly timestamped observations with the expressiveness of the $\nxt$ operator.

Also interesting is to study how to perform AGM style belief revision at the (semantic) level of the equilibria, as in Wang et al~\cite{WangZW13}, though necessarily different since knowledge is not incorporated in the contexts.

Another framework that aims at modeling the dynamics of knowledge is that of evolving logic programs EVOLP~\cite{AlferesBLP02} focusing on updates of generalized logic programs. 
Closely related to EVOLP, hence to \emMCS, are the two frameworks of reactive ASP, one implemented as a solver \emph{clingo}~\cite{GebserGKS11} and one described in~\cite{Brewka13}. The system \emph{clingo} extends an ASP solver for handling external modules provided at runtime by a controller. The output of these external modules can be seen as the observations of EVOLP. Unlike the observations in EVOLP, which can be rules, external modules in \emph{clingo} are restricted to produce atoms so the evolving capabilities are very restricted. On the other hand, \emph{clingo} permits committing to a specific answer-set at each state, a feature that is not part of EVOLP, nor of \emMCS s. Reactive ASP as described in \cite{Brewka13} can be seen as a more straightforward generalization of EVOLP where operations other than $assert$ for self-updating a program are permitted. 
Whereas EVOLP employs an update predicate that is similar in spirit to the $next$ predicate of \emMCS s, it does not deal with distributed heterogeneous knowledge, neither do both versions of Reactive ASP. Moreover, no notion of minimal change is studied for these frameworks.

The dynamics of \emMCS s is one kind of dynamics, but surely not the only one. Studying the dynamics of the bridge rules is also a relevant, non-trivial topic, to a great extent orthogonal to the current development, which nevertheless requires investigation. 

Another important issue open for future work is a more fine-grained characterization of updating bridge rules (and knowledge bases) as studied in \cite{GKL-Clima14} in light of the encountered difficulties when updating rules \cite{SlotaL10,SlotaL12a,SlotaL14} and the combination of updates over various formalisms \cite{SlotaL12a,SlotaL12b}.

Also, as already outlined in \cite{theOtherReactKnowPaper}, we can consider the generalization of the notions of minimal and grounded equilibria \cite{BrewkaE07} to \emMCS s to avoid, e.g., self-supporting cycles introduced by bridge rules, or the use of preferences to deal with several evolving equilibria an \emMCS\ can have for the same observation sequence.

Also interesting is to apply the ideas in this paper to study the dynamics of frameworks closely related to MCSs, such as those in~\cite{HybridJLC13,GoncalvesA10}.

\ack We would like to thank the referees for their comments, which helped improve this paper considerably.
Matthias Knorr and Jo{\~a}o Leite were partially supported by FCT under project ``ERRO -- Efficient Reasoning with Rules and Ontologies'' ({PTDC}/{EIA}-{CCO}/{121823}/{2010}).
Ricardo Gon\c{c}alves was supported by FCT grant SFRH/BPD/47245/2008 and Matthias Knorr was also partially supported by FCT grant SFRH/BPD/86970/2012.

\end{document}